\documentclass[letterpaper, 10 pt, conference]{ieeeconf} 

\IEEEoverridecommandlockouts                        

                            \usepackage{tikz}
\usetikzlibrary{shapes.geometric, positioning, calc}
\usepackage{graphics}
\usepackage{amsmath} 
\usepackage{amssymb}
\usepackage{cite}
\usepackage{tikz}
\usepackage{amsmath}
\usepackage{tabularx}
\usepackage{array}
\usepackage{booktabs}
\usepackage{bm}
\usepackage{algorithm}
\usepackage{algorithmic}
\usepackage{bbm}
\usepackage{caption}
\PassOptionsToPackage{hyphens}{url}\usepackage{hyperref}

\usepackage{tablefootnote}

\definecolor{green}{RGB}{11,155,13}

\title{\LARGE \bf
Toward Human-Like Social Robot Navigation: A Large-Scale, Multi-Modal, Social Human Navigation Dataset
}

\author{Duc M. Nguyen, Mohammad Nazeri, Amirreza Payandeh, Aniket Datar, and Xuesu Xiao
\thanks{All authors are with the Department of Computer Science, George Mason University {\tt\scriptsize \{mnguy21, mnazerir, apayande, adatar, xiao\}@gmu.edu}}
}

\begin{document}
\maketitle
\thispagestyle{empty}
\pagestyle{empty}

\begin{abstract}
Humans are well-adept at navigating public spaces shared with others, where current autonomous mobile robots still struggle:
while safely and efficiently reaching their goals, humans communicate their intentions and conform to unwritten social norms on a daily basis; 
conversely, robots become clumsy in those daily social scenarios, getting stuck in dense crowds, surprising nearby pedestrians, or even causing collisions. 
While recent research on robot learning has shown promises in data-driven social robot navigation, good-quality training data is still difficult to acquire through either trial and error or expert demonstrations. 
In this work, we propose to utilize the body of rich, widely available, social human navigation data in many natural human-inhabited public spaces for robots to learn similar, human-like, socially compliant navigation behaviors. 
To be specific, we design an open-source egocentric data collection sensor suite wearable by walking humans to provide multi-modal robot perception data; we collect a large-scale ($\sim$100 km, 20 hours, 300 trials, 13 humans) dataset in a variety of public spaces which contain numerous natural social navigation interactions; we analyze our dataset, demonstrate its usability, and point out future research directions and use cases.\footnote{Website: \url{https://cs.gmu.edu/~xiao/Research/MuSoHu/}} 
\end{abstract}

\section{INTRODUCTION}
\label{sec::introduction}

Social navigation is the capability of an autonomous agent to navigate in a way such that it not only moves toward its goal but also takes other agents' objective into consideration. Most humans are proficient at such a task, smoothly navigating many public spaces shared with others on a daily basis: humans form lanes or groups among crowds, use gaze, head movement, and body posture to communicate navigation intentions, wait in line to enter congested areas, or give way to others who are in a rush. 
With an increasing amount of autonomous mobile robots being deployed in public spaces \cite{scout, dilligent}, those robots are also expected to navigate among humans in a similar, human-like, socially compliant manner.

\begin{figure}[t]
  \centering
  \includegraphics[width=1\columnwidth]{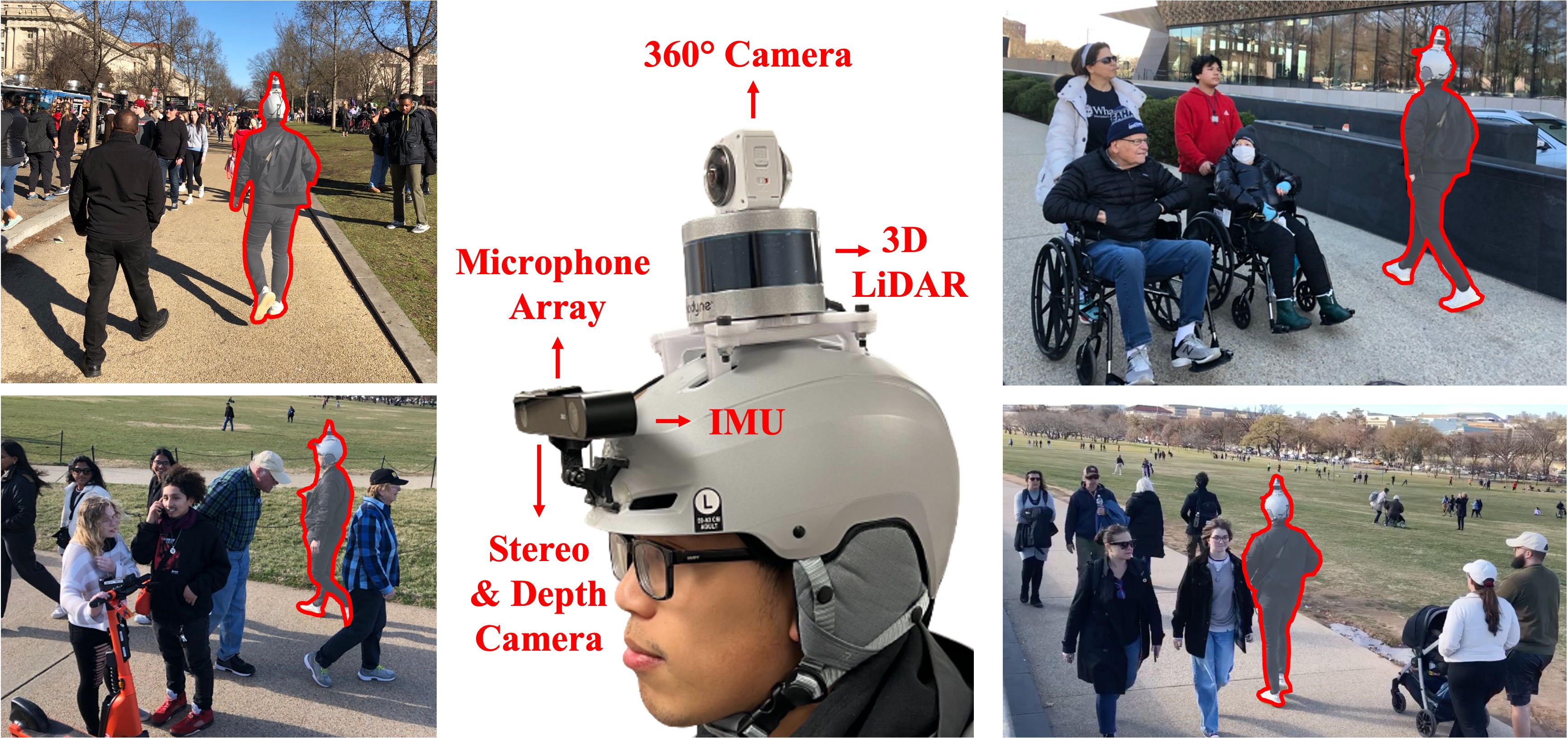}
  \caption{Data collection in natural human-inhabited public spaces with the open-source sensor suite including 3D LiDAR, stereo and depth camera, IMU, microphone array, and 360\textdegree~camera.}
  \label{fig::helmet}
\end{figure}

However, the autonomous navigation performance of these mobile robots is still far from satisfactory. Despite extensive robotics effort to create efficient and collision-free autonomous navigation systems, we still witness the ``frozen robot'' problem in dense crowds and robots moving against upcoming foot traffic or cutting too close to moving humans. Unfortunately, due to such deficiencies, there is increasing fear about the public adoption and even safety of humans around these robots \cite{no_scout, no_starship}. The current lack of safe and socially compliant navigation systems still presents a major hurdle preventing service robots being widely adopted. 

One avenue toward socially compliant robot navigation is using machine learning for robots to learn the variety of unwritten social norms, for which traditional cost functions are hard to design. For example, Reinforcement Learning (RL) \cite{karnan2022voila} uses trial-and-error experiences while Imitation Learning (IL) \cite{tai2018socially} requires expert demonstrations. However, both of these learning paradigms require an extensive amount of training data, which is difficult to acquire: RL in the real world is extremely expensive due to the limited availability of robots, while RL in simulation requires a good model of social navigation interactions of humans, which are what roboticists are trying to create in the first place; IL requires demonstration datasets collected on robot platforms, mostly through expensive human teleoperation at scale \cite{karnan2022scand}. 

Considering the goal of creating socially compliant robot navigation and the availability of many humans that excel at such a task, this work leverages the easily accessible social human navigation data in public spaces for mobile robots to learn from. 
To be specific, we first present an open source, first-person-view, social human navigation data collection sensor suite that can be worn on the head of a walking human and provide easy access to a large body of readily available, high-quality, natural social navigation data in the wild for robot learning, as shown in Fig. \ref{fig::helmet}. Our design includes a set of different robotic sensors: a 3D Light Detection and Ranging (LiDAR) sensor, stereo and depth camera, Inertia Measurement Unit (IMU), microphone array, and 360\textdegree~camera. We open-source our design and software so the sensor suite can be easily replicated and used to collect social human navigation data in different places. 
Second, with the new data collection suite,  we introduce our Multi-modal Social Human Navigation dataset (MuSoHu): a large-scale, egocentric, multi-modal, and context-aware dataset of human demonstrations of social navigation. At the point when this paper is written, MuSoHu contains approximately 20 hours, 300 trajectories, 100 kilometers of socially compliant navigation demonstrations collected by 13 human demonstrators that comprise multi-modal data streams from different sensors, in both indoor and outdoor environments---within the George Mason University campus and the Washington DC metropolitan area. We also provide annotations of interesting social interaction events and of the navigation contexts (i.e., ``casual'', ``neutral'', and ``rush'') for each of the trials.  Third, we present analysis in terms of human and robot social navigation and point out future research directions and anticipated use cases of our dataset.

\section{RELATED WORK}
\label{sec::related_work}
In this section, we review related work in social robot navigation and learning from human datasets. 

\subsection{Social Robot Navigation}

To organically integrate service robots into the fabric of our society, these robots must be capable of moving in human-inhabited spaces in a socially compliant manner. One difficulty in creating such socially compliant navigation systems is to hand-craft appropriate rules or cost functions to cope with unwritten social norms in public spaces~\cite{mirsky2021prevention}. Therefore, researchers have sought help from machine learning and aimed at \emph{learning} socially compliant navigation behaviors in a data-driven manner~\cite{xiao2022motion, baghaei2022deep}. 

RL has shown success in learning a variety of behaviors from simulated trial-and-error experiences~\cite{chen2017socially, xu2021applr, xu2023benchmarking}. However, the high fidelity of simulated social interactions required by RL for social navigation poses its own challenges and requires a good understanding and  then analytical representation of the unwritten social norms to create such simulated interactions, which is the difficulty in social navigation in the first place. Additionally, the reward function in RL needs to be carefully-designed but can still be brittle~\cite{knox2023reward}. 

To address such issues, IL~\cite{bojarski2016end, Nazeri2021} utilizes expert demonstrations to learn socially compliant navigation behaviors~\cite{xiao2022learning, xiao2020appld}. Kretzschmar et al.~\cite{kretzschmar2016socially} has proposed Inverse Reinforcement Learning (IRL) to learn the reward function from demonstrations for social navigation policies. Behavior Cloning~\cite{bojarski2016end, Nazeri2021} has treated the social navigation problem as supervised learning and regressed to an end-to-end motion policy that maps from perception to actions. However, to facilitate IL, a large corpus of socially compliant navigation demonstration data is essential. For example, the Socially Compliant Navigation Dataset (\textsc{scand})~\cite{karnan2022scand} is a recent effort to provide social robot navigation behaviors demonstrated by human teleoperation.

\subsection{Learning from Human Datasets}
\textsc{scand}~\cite{karnan2022scand} is a recent dataset that aims at tackling the challenges of socially compliant robot navigation. \textsc{scand} includes socially compliant, human teleoperated robot navigation demonstrations in indoor and outdoor environments on The University of Texas at Austin campus. Using \textsc{scand}, researchers have shown that IL policies can be trained end-to-end for socially-aware global and local planners for robot navigation. However, \textsc{scand} requires a significant amount of cost and effort to set up and deploy the robot platforms in the wild and to collect large-scale human-teleoperated robot navigation demonstrations to cover the plethora of interesting social interactions in public spaces. Furthermore, how people react differently to a teleoperated mobile robot followed by a human operator is also unclear. 

Considering the difficulty in acquiring large-scale real-world data, researchers have also looked into utilizing recorded videos of human activities in the wild. For example Ego4D~\cite{grauman2022ego4d} is an egocentric video dataset, which offers daily-life activity video of different scenarios (house-hold, outdoor, workplace, leisure, etc.) captured by different humans wearing cameras from different locations worldwide. 
Ego4D offers a solution to the scalability of datasets by introducing a standard and wearable design so many people can collect data in real-world, daily settings from different parts of the world. However, Ego4D is not specifically designed for robotics (hence the lack of common robot sensors and perception like LiDAR, depth camera, IMU, and odometry), so it is difficult for mobile robots to directly learn socially compliant navigation behaviors from the raw video feed in Ego4D.

Inspired by the pros and cons of both \textsc{scand} and Ego4D, we introduce a wearable data collection sensor suite specifically designed to provide data to enable social robot navigation. It allows us to collect social human navigation data from the perspective of a suite of multi-modal robotic sensors in our daily life with a small setup overhead (i.e., with a wearable helmet). We provide a large-scale social human navigation dataset, which can be easily extended in the future by robotics researchers all around the world, show that human-like social robot navigation behaviors can be learned through such a dataset, and point out future research directions and anticipated use cases of our dataset. For other robot navigation datasets which are less relevant to our work compared to \textsc{scand} and Ego4D, we refer the readers to Table I in the \textsc{scand} paper \cite{karnan2022scand}. 
\section{SENSOR SUITE}
\label{sec::hardware}
We design and make publicly available a data collection device, which is wearable by a human walking in public spaces and provides multi-modal perceptual streams that are commonly available on mobile robot platforms.\footnote{\scriptsize\url{https://github.com/RobotiXX/MuSoHu-data-collection}} We also process the raw data to extract human navigation behaviors, i.e., the paths and actions taken by the human demonstrator to navigate through social spaces. 

To be specific, our data collection sensor suite is equipped with a 3D LiDAR, a stereo and depth camera with built-in IMU, a microphone array to provide ambient sound, and a 360\textdegree~camera that offers spherical view of the environment. All the sensors are mounted to a helmet via open-sourced hardware to capture egocentric data of the demonstrator during social navigation. To stream and store real-time social human navigation data, all sensors are connected to a laptop carried by the demonstrator with wired connections (Fig. \ref{fig::helmet} middle). 

\paragraph{3D LiDAR}
As most mobile robots use LiDARs as a reliable sensor to acquire accurate and robust geometric information about the environment, we include a 3D LiDAR to capture such information around the human demonstrator. Considering the different heights of the mounting locations (on robot vs. on our helmet), we use a 3D LiDAR to collect 3D point clouds, which can be converted to 2D scans at different heights if necessary. 
We choose a Velodyne Puck VLP-16 for our sensor suite, which has a range of 100 meters and generates up to 600,000 points/second, across a 360\textdegree~horizontal and 30\textdegree~vertical field of view. The 3D LiDAR is mounted on the top of the helmet to record spatial measurements of the surrounding. 

\paragraph{Stereo and Depth Camera}
RGB cameras provide visual and semantic information of the environment. In addition to the geometric information provided by the LiDAR, semantics also plays a vital role in social navigation interactions. For example, humans use gesture, gaze, and body posture to explicitly or implicitly convey navigational intentions and facilitate interactions. Those behaviors can be used to understand the intentions of other people sharing the same space but are difficult to capture with 3D LiDAR alone. For our sensor suite, we choose Stereolabs ZED 2, a stereo camera with depth sensing and a built-in IMU (see below for more details), considering its compact form factor and efficient power consumption (in contrast to other RGB-D cameras that require a separate power supply, ZED 2 can be efficiently powered by the same USB cable for data transmission). The camera is positioned in the front of the helmet, with the optical axis pointing forward. The wide 120° field of view captures interesting social interactions happening in front of and from the sides of the human demonstrator. 

\paragraph{IMU} 
Many mobile robots are also equipped with IMUs to measure linear accelerations and rotational speeds. 
Therefore, we also utilize the built-in IMU from the ZED 2 camera and record their raw measurements. 
It is worth to note that due to the difference between walking humans and wheeled or tracked robots that drive, the IMU readings collected in our dataset may be significantly different than those from such types of mobile robots, especially the acceleration along the vertical axis. 
We posit that to leverage the IMU data in MuSoHu, special techniques such as transfer learning \cite{weiss2016survey} may be necessary. 

\paragraph{Odometry / Actions}
Similar to \textsc{scand}, we collect visual-inertia odometry provided by the ZED 2 camera. Such positional odometry provides learning data of navigation path and can be utilized to learn robot global planners. 
Different than \textsc{scand}, in which the robot navigation actions can be directly recorded as teleoperation commands, our data collection hardware does not have access to such actions, i.e., how the human demonstrator walks. Therefore, we extract linear and angular velocities from the positional odometry using the difference between two consecutive odometry frames.  

\paragraph{360\textdegree~Camera}
In addition to the forward facing stereo and depth camera, we also collect 360\textdegree~RGB video to provide better situational awareness of the surrounding and include all possible sensory information that can be
provided by active pan-tilt cameras onboard many mobile robot platforms. 
We use a Kodak Pixpro Orbit360 4K VR Camera to collect 360\textdegree~images. The camera has a very compact form factor with two lenses integrated in one camera body to provide spherical 360\textdegree~view. Note that due to software limitations the camera's webcam mode does not allow both lenses to stream live video to the laptop, so we save the spherical 360\textdegree~view from both lenses to an SD card in the camera. 

\paragraph{Microphone Array}
Although not commonly used for navigation tasks, microphones are available on many mobile robot platforms, e.g., for verbal communications. Furthermore, recent research has started to investigate using sound for navigation \cite{chen2021structure}. Considering the extra information provided by this different perception modality, we also include a microphone array, a Seeed Studio ReSpeaker Mic Array v2.0, to collect ambient sound during social human navigation. 

\section{DATASET}
\label{sec::dataset}

\begin{figure*}
  \centering
  \includegraphics[width=1\textwidth]{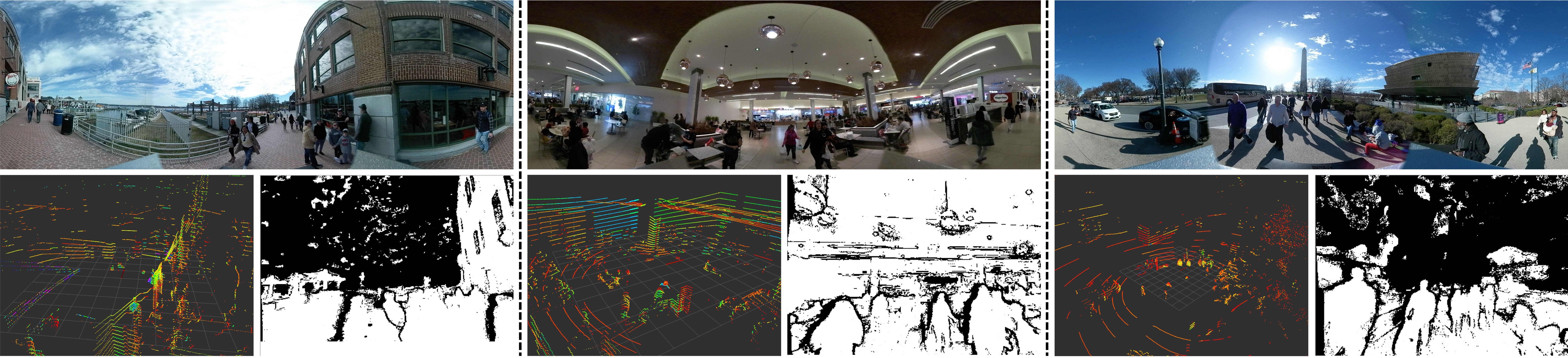}
  \caption{Three example data frames in Old Town Alexandria, VA, Springfield Towncenter, VA, and National Mall, Washington DC. 360\textdegree~ view (top), 3D LiDAR point cloud (bottom left), and depth image (bottom right) are shown for each data frame. }
  \label{fig::scenes}
  \vspace{-10pt}
\end{figure*}

The sensor suite described in Sec. \ref{sec::hardware} is designed to be easily replicable by any research group and to collect data worldwide. But we collect an initial Multi-modal Social Human navigation dataset (MuSoHu) on the George Mason University campus and in the Washington DC metropolitan area (Fig. \ref{fig::scenes}).\footnote{\url{https://dataverse.orc.gmu.edu/dataset.xhtml?persistentId=doi:10.13021/orc2020/HZI4LJ}}

\subsection{Data Collection Procedure}
To collect multi-modal, socially compliant, human-level navigation demonstrations to learn future robot navigation, seven human demonstrators wear the sensor suite helmet and navigate to predefined goals in public spaces in a socially compliant manner. We choose navigation scenarios with frequent social interactions in various indoor and outdoor environments at different time periods (e.g., after class or during weekends). The sensor suite's superior portability (i.e., only a helmet and a laptop) also allows us to record portions of MuSoHu in other settings in the Washington DC Metropolitan Area, including Fairfax, Arlington, and Springfield in Virginia and the National Mall in DC. 
Notably, for a trajectory at a certain location at the same time period, in many cases, we record three trials to capture three navigation contexts, i.e., \emph{casual}, \emph{neutral}, and \emph{rush}, in which walking speed and safety distance from others may vary, in order to encourage different social navigation interactions based on different contexts. We intend such context awareness in MuSoHu to be useful for future studies on context-aware social navigation, e.g., social compliance when facing someone who is about to be late for a class is different than that when facing someone who is taking a casual afternoon stroll in the park. 

For each trajectory, all sensor data are collected using the Robot Operating System (\textsc{ros}) Bag functionality, except the 360\textdegree~camera, which does not allow data streaming of both built-in cameras to provide spherical 360\textdegree~view to \textsc{ros}. Therefore, we store the 360\textdegree~video on an SD card and provide synchronization using a movie clapboard.

\subsection{Dataset Analyses}
\subsubsection{Labeled Annotations of Social Interactions}
MuSoHu includes a list of textual tags for each trajectory that describe the different social interactions that occur along the path. 
We expand beyond the tags from \textsc{scand} and the full list of 17 predefined labels can be found in Table \ref{tag_table} (with five new tags in bold font). 

\begin{table}[!h]
\centering
\caption{Descriptions of Label Tags  Contained in MuSoHu.}
\begin{tabular}{>{\centering\arraybackslash}m{0.1\textwidth}>{\centering\arraybackslash}m{0.25\textwidth}>{\centering\arraybackslash}m{0.05\textwidth}}
\toprule
 \textbf{Tag} & \textbf{Description} & \textbf{\# Tags}\\
 \toprule
 Against Traffic & Navigating against oncoming traffic & 210\\ 
 \midrule
 With Traffic & Navigating with oncoming traffic & 170\\
 \midrule
 Street Crossing & Crossing across a street & 120\\
 \midrule
 Overtaking & Overtaking a person or groups of people  & 100\\
 \midrule
  Sidewalk & Navigating on a sidewalk  & 160\\
  \midrule
  Passing Conversational Groups & Navigating past a group of 2 or more people that are talking amongst themselves  & 94\\
  \midrule
  Blind Corner & Navigating past a corner where the human cannot see the other side & 90\\
  \midrule
  Narrow Doorway & Navigating through a doorway where the human opens or waits for others to open the door & 45\\
  \midrule
  Crossing Stationary Queue & Walking across a line of people  & 50\\
  \midrule
  Stairs & Walking up and/or down stairs & 30\\
  \midrule
  Vehicle Interaction & Navigating around a vehicle & 26\\
  \midrule
 Navigating Through Large Crowds & Navigating among large unstructured crowds & 45\\ 
 \midrule
 \textbf{Elevator Ride} & Navigating to, waiting inside, and exiting an elevator & 15\\ 
  \midrule
 \textbf{Escalator Ride} & Navigating to and riding an escalator & 6\\ 
  \midrule
 \textbf{Waiting in Line} & Waiting in Line to enter congested areas & 5\\ 
  \midrule
 \textbf{Time: Day} & Navigation during day time & 150\\ 
  \midrule
\textbf{Time: Night} & Navigation during night time & 40\\ 
\bottomrule
\end{tabular}
\label{tag_table} 
\vspace{-10pt}
\end{table}

\subsubsection{Example Data Frames}
In Fig. \ref{fig::example}, we show the corresponding linear and angular velocities (filtered by Savitzky-Golay filter to smooth out high frequency noises caused by walking gait) and navigation path taken by the human demonstrator in the three scenarios shown in Fig. \ref{fig::scenes}. In the first scenario, the demonstrator navigates around a right corner and avoids an upcoming family; in the second scenario, the demonstrator makes a 90\textdegree~right-hand turn, while avoiding people in an indoor food court; in the third scenario, the demonstrator dodges (right-left-right) a dense crowd during a right-hand turn. Both linear and angular velocities and navigation path provide learning signals for mobile robots. 

\definecolor{pathgreen}{rgb}{0.098, 1, 0} 
\begin{figure}[t]
  \centering
  \includegraphics[width=1\columnwidth]{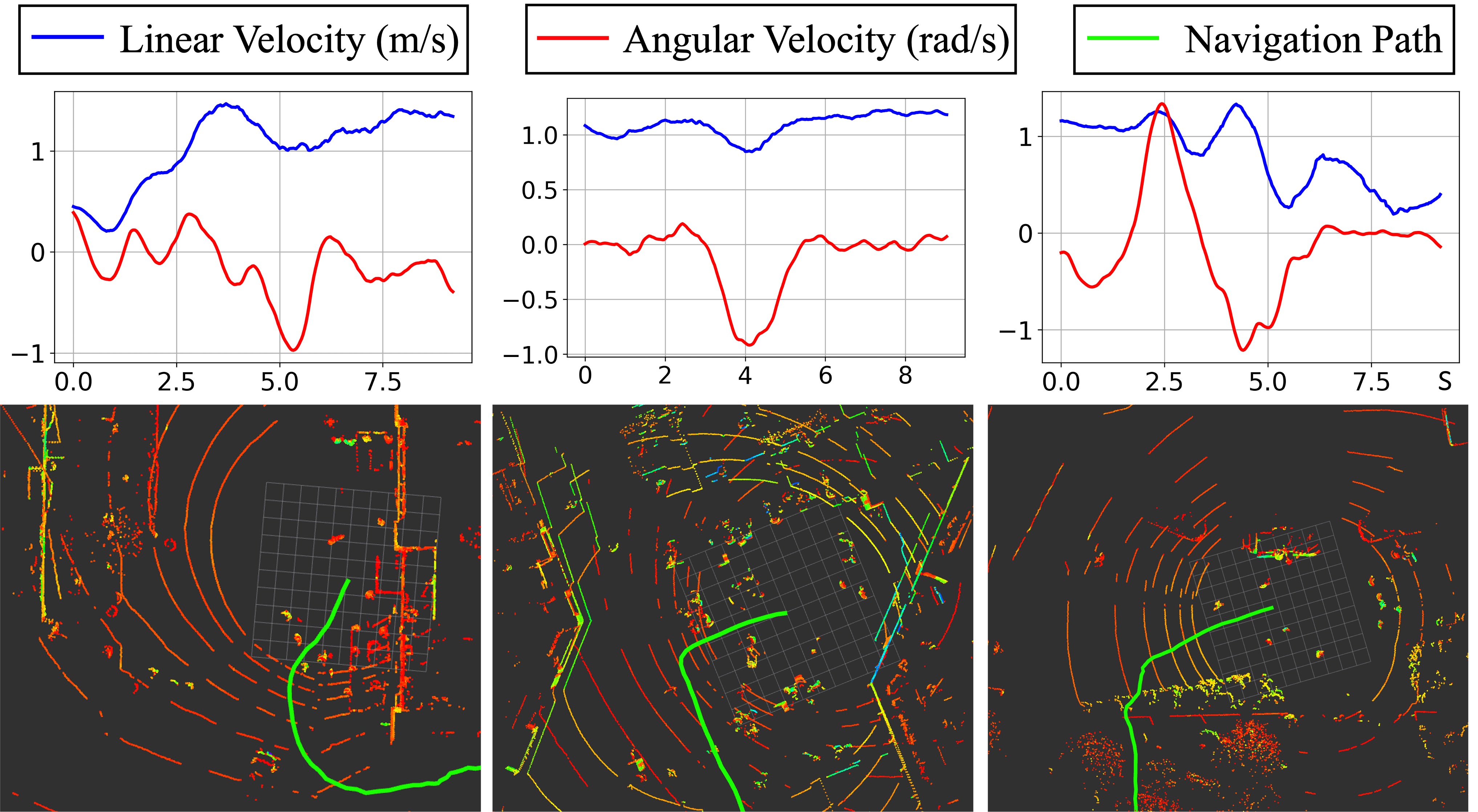}
  \caption{Linear (\textcolor{blue}{Blue}) and Angular (\textcolor{red}{Red}) Velocities and Navigation Path (\textcolor{pathgreen}{Green}) Taken by the Human Demonstrator. }
  \label{fig::example}
  \vspace{-11pt}
\end{figure}

\subsubsection{Proof-of-Concept Usage}
\begin{figure}[t]
  \centering
  \includegraphics[width=0.8\columnwidth]{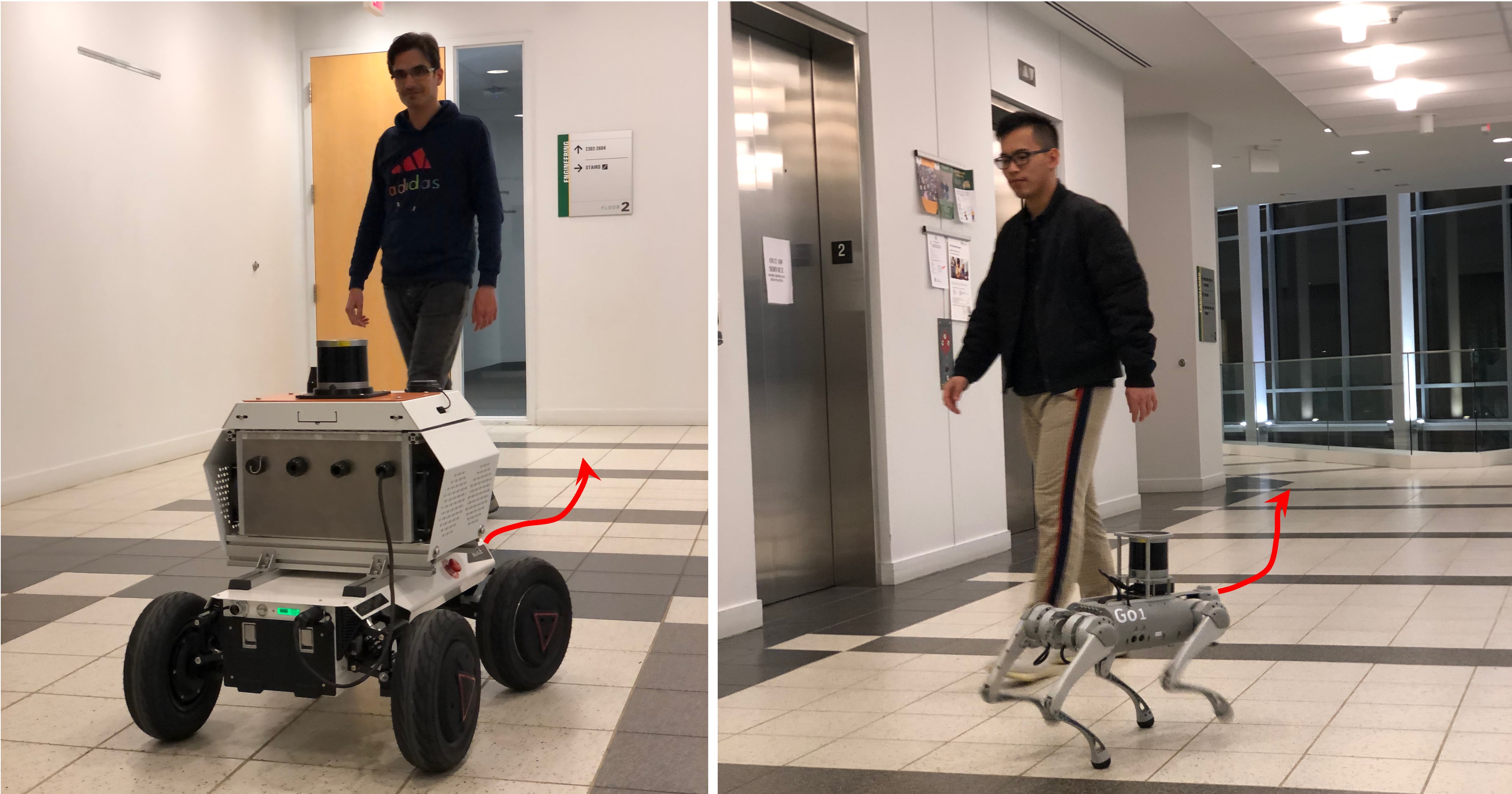}
  \caption{Learned Obstacle Avoidance Behavior from MuSoHu.}
  \label{fig::robots}
\end{figure}

We use a small subset of MuSoHu data (ten navigation trials) to train a Behavior Cloning policy that maps from raw LiDAR input to linear and angular velocity (Fig. \ref{fig::example}). The learned policy is deployed on two physical robots, an AgileX Hunter SE (an Ackermann steering wheeled vehicle) and a Unitree Go1 (a quadruped robot), both of which exhibit collision avoidance behavior learned from MuSoHu (Fig. \ref{fig::robots}). 

\section{ANTICIPATED USE CASES}
\label{sec::analysis}

MuSoHu's large body of socially compliant human navigation data with multi-modal robotic perception collected in natural public spaces in the wild provide opportunities for many future research directions. 

\subsection{Learning Social Robot Navigation}
The primary purpose of MuSoHu is to provide a large corpus of training data for mobile robots to learn socially compliant navigation behaviors. As we demonstrate in our preliminary experiments, robot navigation behaviors similar to the human behaviors in MuSoHu can be learned end-to-end using Behavior Cloning. Other imitation learning methods, such as IRL, can utilize MuSoHu to learn a socially compliant cost function for downstream navigation planners~\cite{kretzschmar2016socially}. 

The replicability of our sensor suite makes collecting social human navigation data very easy. We intend the sensor suite to be replicated by different research groups to collect data in different countries worldwide. An even larger and also more diverse corpus of data opens up orthogonal research directions that are not currently possible. For example, new social robot navigation systems can be developed that are culturally dependent, i.e., the way a mobile robot moves can be fit into different culture contexts. For example, imagine a contact-tolerant culture where pedestrians are comfortable with walking very closely to each other vs. a contact-averse culture where people prefer to keep distance. 

\subsection{Imitation Learning with Various Constraints}
One potential challenge , in other words, opportunity for future research, is how to address the difference in human and robot navigation. 
Human navigation is based on legged locomotion, while most mobile robots are wheeled or tracked. Different motion morphologies caused by such an embodiment mismatch \cite{hudson2022skeletal} may require extra care to be taken during learning. Transfer learning techniques \cite{weiss2016survey} may provide one promising avenue to leverage the full potential of MuSoHu. 
In addition to the different motion morphologies, despite our choice of sensor modalities to align with robot perception, viewpoint mismatch still exists: to avoid occluding the 3D LiDAR view, it is mounted on top of the helmet, which is higher than most robots’ LiDAR position; all perceptual data are subject to the effect of walking gait cycles, e.g., the cyclic motion along the vertical axis, which does not exist for most mobile robots. 
Therefore, imitation learning from mismatched observation techniques \cite{karnan2022voila} need to be investigated to address the perceptual mismatch between MuSoHu and mobile robots. 

\subsection{Studying Social Human and Robot Navigation}
One question being debated frequently is \emph{should roboticists build robots to navigate in public spaces in the same way as humans?} Our MuSoHu dataset, along with its future extensions in different countries worldwide, and \textsc{scand} provide a way to investigate related problems. Assuming the navigation behaviors in MuSoHu and \textsc{scand} are the optimal way of human and robot social navigation in public spaces respectively, we can analyze both datasets to see whether the human and robot behaviors are the same, similar, or completely different. Another way is to build social robot navigation systems with the data in MuSoHu and \textsc{scand} and evaluate the learned social navigation behaviors with standardized protocols and metrics~\cite{pirk2022protocol} to see wheter there is any difference between the two and if yes which way is preferred by people that interact with the robots. 

\subsection{Real-to-Sim Transfer for Social Navigation}
Creating high-fidelity social navigation simulation environments has been a focus of social robot navigation researchers~\cite{tsoi2020sean, holtz2022socialgym}. A realistic simulator that can induce real-world human-robot social interactions that conform with the underlying unwritten social norms will facilitate social robot navigation research on multiple fronts, such as reinforcement learning based on simulated trial and error, large scale validation and evaluation of new social navigation systems before their real-world deployment, and objective and repeatable benchmark and comparison among multiple social navigation systems. However, existing social navigation simulators rely on simplified human-robot interaction models, e.g., the Social Force Model~\cite{helbing1995social} or ORCA~\cite{van2011reciprocal}. Such a sim-to-real gap~\cite{liang2021crowd} may cause problems when the navigation systems learned, evaluated, or compared in simulation are deployed in the real world.

The MuSohu dataset provides another alternative and promising avenue toward shrinking such a sim-to-real gap through real-to-sim transfer to improve social navigation simulation. The data collected in the wild in MuSoHu enable researchers to synthesize natural, real-world, human-robot social navigation interactions in simulation. Approaches can be developed to learn such interaction models from the natural interactions in MuSoHu, which can be used to control simulated agents, robots or humans, in a high-fidelity simulator.

\subsection{Investigating Robot Morphology for Social Navigation}
Human-human social navigation interactions embody a large set of interaction modalities, which are frequently present in MuSoHu. For example, in addition to avoiding other humans as moving obstacles, humans use gaze, head movement, and body posture to communicate navigation intentions in crowded spaces; they use body or natural language to express their navigation mindset or context (e.g., they are in a rush and apologize for being less polite). Most current mobile robots, however, do not possess such capabilities to communicate their navigation intentions and contexts in an efficient manner. Analyzing the human-human social navigation interaction modalities in MuSoHu will shed light on what other robot morphology may be useful to facilitate efficient social navigation, such as adding a robot head with gaze~\cite{hart2020using}, turn signals~\cite{unhelkar2015human}, or gait features (for legged robots)~\cite{kruse2014evaluating} to disambiguate navigation intentions, or adding voice~\cite{medicherla2007human} to signal the urgency of the robot's navigation task. 
\section{CONCLUSIONS}
\label{sec::conclusions}
We present a large-scale, multi-modal, social human navigation dataset, MuSoHu, to allow robots to learn human-like, socially compliant navigation behaviors. Our open-sourced hardware and software design allows our portable sensor suite to be easily replicated and used to collect social human navigation data in a variety of public spaces worldwide. Such an easy access to a variety of natural social navigation interactions in human-inhabited public spaces in the wild is shown in our preliminary experiments to be useful to learn social robot navigation. We point out future anticipated use cases and research directions of MuSoHu to develop socially compliant robot navigation.

\bibliographystyle{IEEEtran}
\bibliography{IEEEabrv,references}
\end{document}